\documentclass[letterpaper, 10 pt, conference]{ieeeconf}

\IEEEoverridecommandlockouts
\usepackage{cite}
\usepackage{amsmath,amssymb,amsfonts}
\usepackage{algorithmic}
\usepackage{graphicx}
\usepackage{subcaption}
\usepackage{multirow}

\usepackage{color}

\usepackage{url}
\usepackage{textcomp}
\usepackage{xcolor}
\usepackage{bm}

\title{\LARGE \bf RICO-MR: An Open-Source Architecture for Robot Intent Communication through Mixed Reality}

\author{Simone Macciò$^{*}$, Mohamad Shaaban, Alessandro Carfì, Renato Zaccaria and Fulvio Mastrogiovanni
\thanks{This research was made, in part, with the Italian government support under the National Recovery and Resilience Plan (NRRP), Mission 4, Component 2 Investment 1.5, funded from the European Union NextGenerationEU and awarded by the Italian Ministry of University and Research and by the NVIDIA Academic Hardware Grant Program.}
\thanks{All the authors are with the Department of Informatics, Bioengineering, Robotics, and Systems Engineering, University of Genoa, Via Opera Pia 13, 16145 Genoa, Italy. 
{\tt\small simone.maccio@edu.unige.it}}%
}

\begin{document}

\onecolumn
© 2023 IEEE.  Personal use of this material is permitted.  Permission from IEEE must be obtained for all other uses, in any current or future media, including reprinting/republishing this material for advertising or promotional purposes, creating new collective works, for resale or redistribution to servers or lists, or reuse of any copyrighted component of this work in other works.
\newpage
\twocolumn
\maketitle
\thispagestyle{empty}
\pagestyle{empty}

\begin{abstract}
This article presents an open-source architecture for conveying robots' intentions to human teammates using Mixed Reality and Head-Mounted Displays. The architecture has been developed focusing on its modularity and re-usability aspects. Both binaries and source code are available, enabling researchers and companies to adopt the proposed architecture as a standalone solution or to integrate it in more comprehensive implementations. Due to its scalability, the proposed architecture can be easily employed to develop shared Mixed Reality experiences involving multiple robots and human teammates in complex collaborative scenarios.
\end{abstract}


\section{Introduction}

Driven by the principles of the fourth industrial revolution, the era of human-centered manufacturing and smart factories has given prominence to Human-Robot Collaboration (HRC) as a research field from both scientific and industrial perspectives. HRC aims to combine the benefits of human cognitive abilities with machine efficiency and speed \cite{villani2018survey} by allowing humans and robots to share physical space and duties \cite{arents2021human}. Therefore, collaborative platforms should consider safety and flexibility principles in their design to allow close interaction with human teammates in a wide range of collaborative tasks \cite{matheson2019human}. Nevertheless, open problems regarding collaboration effectiveness remain unsolved. In particular, there is the theme of reliable communication between agents. Successful collaboration requires a robot to understand and react to the actions of humans, which can take different forms, e.g., voice or gestures \cite{carfi2018online,liu2018deep,bongiovanni2022gestural}, or can be implicitly expressed through nonverbal cues \cite{li2017implicit,lastrico2021movement}. Simultaneously, the human operator requires clear feedback and intuitive media to anticipate the robot's actions. Feedback is fundamental for reducing the cognitive load and avoiding unexpected robot movements that may cause accidents or task failures. Poor communication can lead to undesired outcomes, undermining the quality of collaboration and the trust of humans in the robot. 

\begin{figure}[t!]
\centering
\begin{subfigure}{.48\textwidth}
  \centering
  \includegraphics[width=\linewidth]{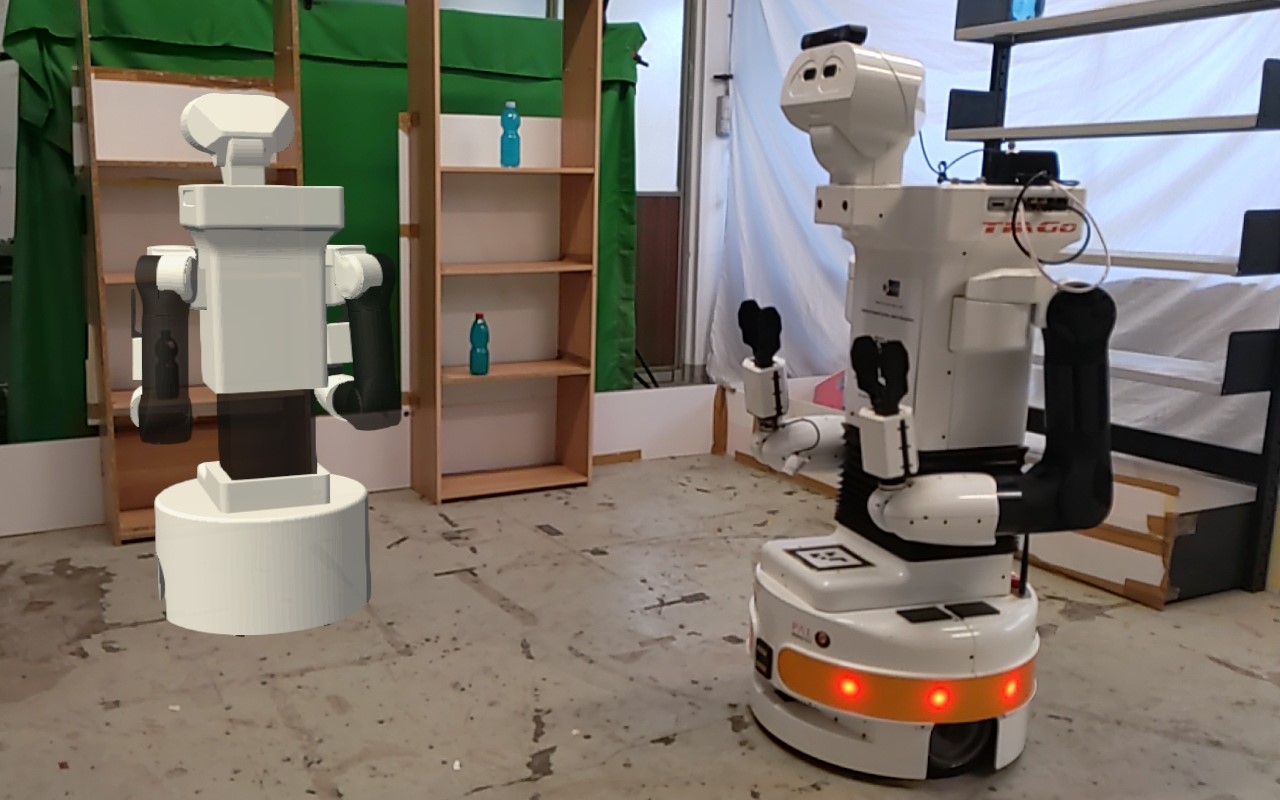}  
  \caption{Display of navigation intentions.}
  \label{fig:navigation-intention}
\end{subfigure}
\par\medskip
\begin{subfigure}{.48\textwidth}
  \centering
  \includegraphics[width=\linewidth]{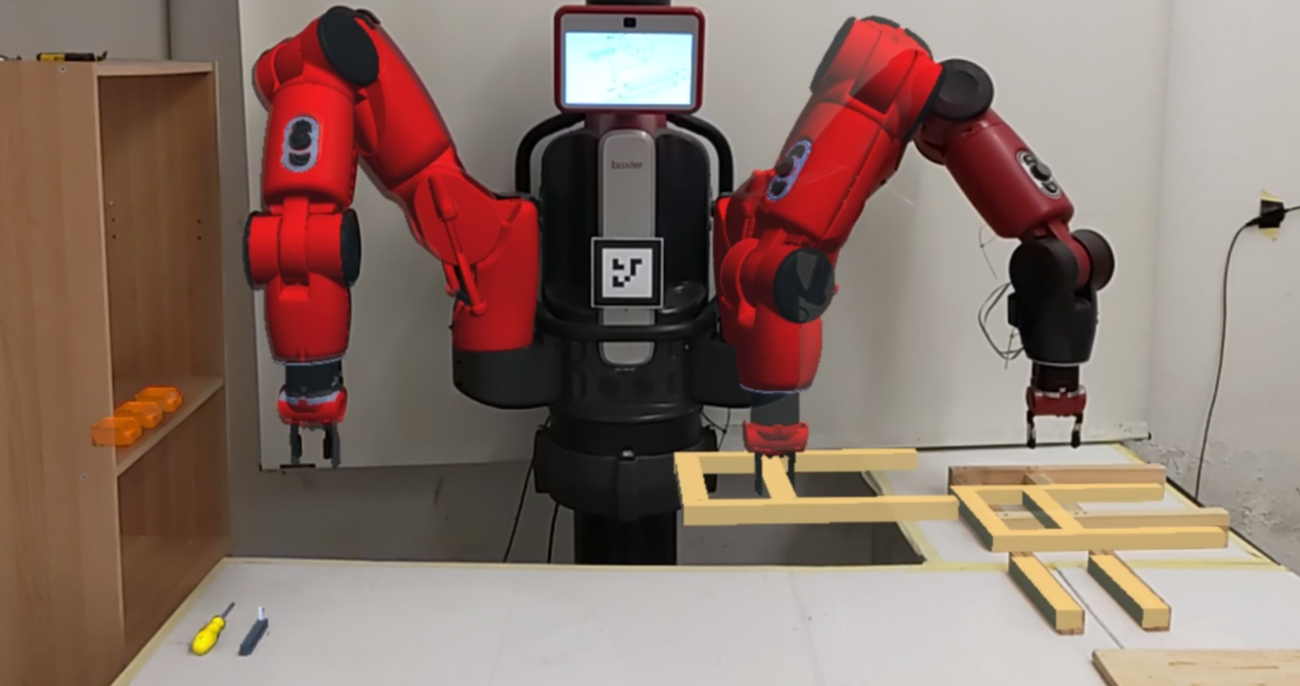}  
  \caption{Display of manipulation intentions.}
  \label{fig:manipulation-intention}
\end{subfigure}
\caption{
Visual examples of mixed reality solutions for conveying robots' intentions to human partners. Specifically, the pictures show the first-person view of a user wearing a head-mounted display and experiencing holographic communication. In Fig. \ref{fig:navigation-intention}, a mobile robot conveys its navigation intentions by displaying a dynamic hologram moving along the planned trajectory. Similarly, in Fig. \ref{fig:manipulation-intention}, the robotic platform previews an upcoming manipulation action during a collaborative assembly task.
}
\label{fig:anticipatory-modalities}
\end{figure}

\begin{figure*}[t!]
    \centering
    \includegraphics[width=0.95\textwidth]{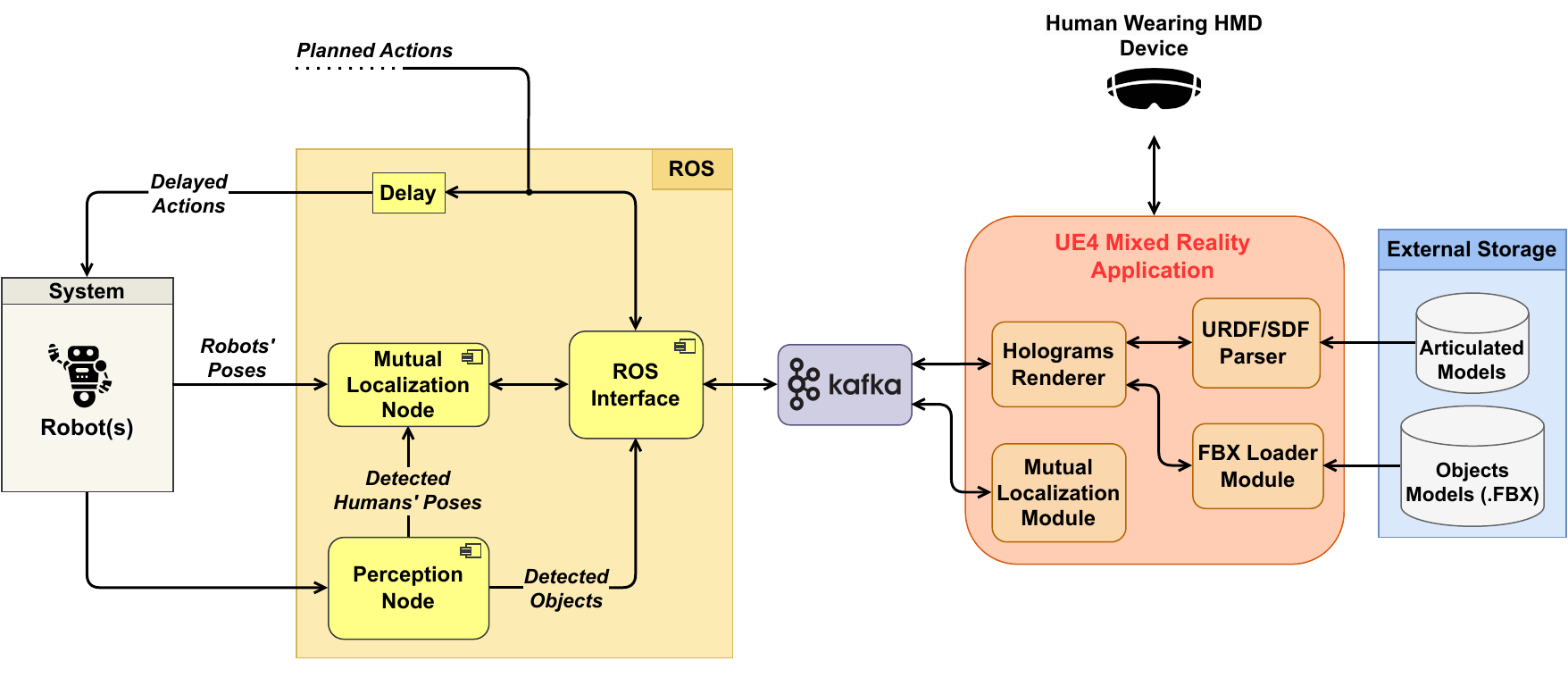}
    \caption{Overview of the \textit{RICO-MR} architecture.}
    \label{fig:software-architecture}
\end{figure*}

Observing human-human collaboration, it is evident that interpreting implicit signals and consequently associating them with the intent of the co-worker is an essential ability. Several studies \cite{klein2005common, mutlu2009nonverbal} have reported that humans seamlessly collaborate by communicating and inferring each other's intentions via implicit cues, including gaze, posture, and gestures. Understanding teammates' intentions is crucial in collaborative tasks, especially when concurrent activities are required, since it ensures agents' ability to synchronize actions, thus maximizing team efficiency. Consequently, seamless human-robot collaboration can be achieved only if humans can understand robot teammates' intentions online naturally and straightforwardly. Nevertheless, conveying a robot's intentions in human-robot teams is challenging, especially in industrial settings where robotic platforms lack humanoid features and cannot replicate natural implicit cues such as gaze. Numerous alternatives have been proposed to this end. Vocal \cite{nikolaidis2018planning} interfaces have been explored; however, they are easily affected by background noise, which is abundant in industrial settings. Several visual strategies have also been proposed, ranging from intent-carrying blinking interfaces \cite{8520655} to external displays used to show upcoming robot actions \cite{shrestha2016intent}, up to projection systems used to highlight objects or areas of interest directly on the collaborative workspace \cite{andersen2016projecting,vogel2017safeguarding}. However, these forms of visual feedback may appear confusing and counterintuitive to human teammates, requiring an increased cognitive load on their side. Furthermore, these approaches share the need for a structured environment that limits the spectrum of applications in industrial settings. 

Recently, Augmented Reality (AR) through handheld devices has been adopted to overlay holograms to the environment, allowing an additional communication modality at the expense of operator autonomy \cite{chacko2019augmented}.
However, with the introduction of Optical See-Through Head-Mounted Display (OST-HMD) devices, like the \textit{Microsoft HoloLens} family, the concept of Mixed Reality (MR) started emerging, enabling the creation of hybrid experiences where users perceive holograms contextualized to the surrounding environment, all the while maintaining the hands-free to interact with the workspace. To this extent, recent comprehensive reviews \cite{dianatfar2021review, suzuki2022augmented} report how such devices can be employed in industrial HRC scenarios to achieve safer interactions, either through the definition of holographic safety boundaries \cite{hoang2022virtual} or by projecting the robot's workspace \cite{ostanin2020human} to improve operator's awareness. In such a context, MR can be employed not only for safety purposes but also to improve human-robot communication, broadening robots' expressiveness and enabling them to convey their upcoming intentions to human teammates as holographic cues. Several studies \cite{rosen2019communicating, newbury2022visualizing, maccio2022mixed, hoang2022virtual} have, in fact, demonstrated how MR could be a suitable communication medium in HRC, providing humans with a better understanding of a robot's internal reasoning and leading to more effective collaboration.  
Nevertheless, although multiple studies have assessed the benefits of MR in industrial HRC, there are yet no available open-source software solutions that can aid researchers and companies in taking advantage of such a holographic communication scheme. Therefore, we plan to bridge the gap in the field by introducing \textit{RICO-MR}, an open-source architecture for Robot Intent Communication through MR, which unlocks a new communication modality that can be adopted in whichever robotic scenario.



\section{Software Architecture} \label{sec:softarchitecture}

The proposed \textit{RICO-MR} software architecture is illustrated in Fig. \ref{fig:software-architecture}. It comprises two macro components: the \textit{Mixed Reality Application}  (on the right), built with Unreal Engine 4.27 (UE4), and the \textit{System's Architecture} (on the left). The communication between the two components operates through \textit{Apache Kafka}, an open-source, high-performance distributed streaming platform.

\subsection{Mixed Reality Application} \label{section:mr-app}

The MR application plays a central role in the architecture and is deployed on an HMD device worn by a human teammate. It is responsible for rendering the holographic layer used by the robot to convey its intentions. Within its UE4 implementation, the holographic layer is built using Microsoft's \textit{Mixed Reality UX Tools} plugin, a popular framework providing building blocks and functionalities to develop 3D virtual experiences and targeting a specific family of HMDs, namely Microsoft HoloLens, and HoloLens2. Moreover, the UE4 application possesses modules that are capable of parsing and loading 3D models at runtime. Both simple and articulated models can be employed to ensure that holograms of robots, items, and tools relevant to the collaboration can be spawned inside the MR layer. This feature, in turn, implies that the MR interface can convey complex robot intentions involving, for example, an agent interacting with objects in a shared workspace, offering additional insight to the human teammate.
In particular, a simple side menu allows users to select which robot model to load inside the MR scene. As we will later detail in Section \ref{sec:example-of-usage}, the MR application ships with several pre-loaded models. However, the list can be extended by specifying an appropriate remote storage repository in the application settings. Such a repository can be employed to store relevant robot resources (i.e., URDF and SDF files) and is used at run-time to refresh the list of models ready to be spawned as holographic assets.

Furthermore, the application provides tools for mutual localization between the HMD and the rest of the system. Specifically, we used the built-in functionalities of Microsoft's MR plugin, which offers QR code tracking capabilities. To this extent, a simple QR code is employed to establish the initial mutual localization between the HMD and the robot in the real world, ensuring that the holographic model is spawned consistently with its counterpart. Upon completing this initial phase, the HMD continuously publishes its updated pose (i.e., position and orientation in space), ensuring consistency between the mutual localization of the human and robot, even if the user moves around the environment. 
Although the employed plugin can only track one QR code at a time, the application can easily accommodate multi-robot scenarios. The underlying implementation stores the data payload of each tracked marker. As such, it is possible to spawn several robot models inside the MR layer by simply generating as many QR codes with different textual content. However, users need to keep in mind that the mutual localization phase is carried out only once, with the first QR code tracked, which effectively plays the role of spatial anchor between HMD and surrounding environment. Therefore, regardless of the number of markers employed in the particular scenario, the HMD's pose is always updated and published with respect to such anchor. Nevertheless, it is always possible to query the system to find out relative localization between the first QR code and subsequent ones, ensuring that consistent spatial relationship between HMD and markers can be computed at any time.

Finally, coherently with how custom robot resources are handled, the application also ships with a link to a repository containing FBX files of objects of common use which can be loaded as holograms on \textit{system}'s request. The link can be modified in the application settings, enabling users to customize the holographic layer according to their particular HRC scenario. This feature also ensures that users can effectively adopt the application off-the-shelf, with no need to compile the project and manually load their FBX files inside the UE editor. Nevertheless, to provide support to researchers and companies interested in expanding the capabilities of the architecture or adding modules to it, we decided to make the UE4 project publicly available under MIT license\footnote{\url{https://github.com/TheEngineRoom-UniGe/RICO-MR}}.

\subsection{System's Architecture} \label{section:system-arch}

The \textit{system} is purposely denoted by general terms to indicate that the proposed architecture is independent of the adopted robotic platform. Its intrinsic modularity, combined with the QR code tracking capabilities, allows scaling up to accommodate multiple robots simultaneously, thus broadening the range of human-robot interactive scenarios in which the architecture can be employed. Moreover, the system can account for external sources of data (e.g., motion capture systems and external depth cameras), which can be integrated into ad hoc applications. In the context of this work, the system's architecture is implemented using the Robot Operating System (ROS) \cite{quigley2009ros}, a popular middleware for developing robotics applications. However, as later discussed when detailing the role of Kafka in the architecture, the ROS adoption is optional and can be partially or fully replaced. 

The system provides \textit{perception} and \textit{localization} capabilities. Both can originate directly from the robot(s) or external sources. Through perception, the robot(s) perceive the tools and objects inside the collaborative space, recognize them through appropriate object detection models, and track their poses. This information flows through Kafka to the MR application, which loads the corresponding resources (i.e., FBX files) and spawns the holograms of the objects consistently with the real world, as can be observed in Fig. \ref{fig:manipulation-intention}. Once objects are detected by the perception and their holograms are spawned in the MR app, such holographic counterparts can be used by the robot to project its intentions to interact with relevant items. As an example, Fig. \ref{fig:manipulation-intention} displays a robot's upcoming pick-and-place action during a collaborative assembly process, providing intuitive feedback to the human colleague about the next piece the robot intends to manipulate. 

The system's perception can also account for detecting and tracking the human teammate's pose, which is fed to the localization node. This component maintains a coherent estimate of the mutual localization between agents by merging the perception data with the localization data from the HMD. This mechanism ensures that the holographic representations are consistent even in scenarios of mobile collaboration, where both human and robotic agents move throughout the shared workspace.

\subsection{Conveying Robot's Intentions}

Regarding the robot's actions to be displayed as holographic intentions, the proposed architecture is modular, enabling developers to integrate their custom action planners. As such, it is possible to deal with manipulation and/or navigation actions, adapting the architecture to any collaborative scenario. 
As an example, for ROS-based solutions, the architecture can easily be interfaced respectively with \textit{MoveIt} for motion planning and manipulations and with \textit{Navigation Stack} for path planning purposes.

More in detail, a list of Kafka topics exposed by the architecture is presented in Table \ref{table:topics}. For each robot model tracked within the holographic layer, two topics are available, respectively the \textit{/robot/\{id\}/navigation\_plan} for previewing navigation actions and \textit{/robot/\{id\}/joint\_trajectory} for manipulation actions. The \textit{id} parameter is automatically assigned by the architecture once a new robot model is identified through the QR code and its holographic representation spawned in the MR scene. This mechanism allows the architecture to manage multiple robots at once.

\renewcommand{\arraystretch}{1.3}
\begin{table}[h!]
 \caption{
List of Kafka topics available to interact with the architecture. We report, for each topic, the corresponding ROS message type to ensure support for ROS-based solutions. In particular, conversion from the ROS message to its equivalent representation in Kafka could be achieved by filling the Kafka message's payload with the JSON-serialized content of the ROS message.}
\resizebox{0.48\textwidth}{!}{%
\centering
 \begin{tabular}{ |l|l| }
  \hline
  \multicolumn{2}{|c|}{\textbf{Published topic}} \\
  \hline
  \textbf{Topic name} & \textbf{ROS message type} \\
  \hline
  /hmd/\{id\}/pose & \textit{geometry\_msgs/PoseStamped} \\
  \hline
  \multicolumn{2}{|c|}{\textbf{Subscribed topics}} \\
  \hline
  \textbf{Topic name} & \textbf{ROS message type} \\
  \hline
  /robot/\{id\}/navigation\_plan & \textit{nav\_msgs/Path} \\
  /robot/\{id\}/joint\_trajectory & \textit{trajectory\_msgs/JointTrajectory} \\
  /object/\{id\}/state & \textit{rico\_msgs/state} \\
  \hline
\end{tabular}}
 \label{table:topics}
\end{table}

Overall, planned robot actions published on the respective topics are dispatched through Kafka and received by the MR app, which proceeds to render them as holographic animations and, consequently, ensures dynamic previews of the robot's intentions.
At the same time, a small, fully parameterizable \textit{delay} is introduced to the real action executed on the robot to ensure that the human teammate can experience the holographic action first. 
During previous experimental validations carried out in \cite{maccio2022mixed}, the delay has been set to $3$ \textit{seconds}, an empirical value which proved a reasonable trade-off between real-time performances and communication capabilities, ensuring holographic anticipation of upcoming robot's actions without slowing the collaborative pace noticeably.

Table \ref{table:topics} also shows a \textit{/object/\{id\}/state} topic, which is exposed by the architecture and purposely intended for perception components being used within the particular HRC scenario. This topic makes it possible to integrate external perception pipelines, enabling robot(s) to detect and track objects in the collaborative space and spawn the corresponding holographic representations accordingly. To this extent, the \textit{id} parameter can be configured so that each tracked object has its own Kafka topic.
Additionally, the topic's type is custom-defined, enabling developers to specify the object category (as a string field), its pose, and its joint state in the case of articulated objects. Such information is received by the MR app, which proceeds to load the resources and spawn the holographic model consistently. The ROS package, including definitions of custom message types, is provided in the accompanying repository.

Finally, as mentioned in the previous paragraphs, the architecture publishes the HMD's pose tracking in a topic named \textit{/hmd/\{id\}/pose}. This information is made available to the external world for third-party applications needing localization information to develop multiuser or shared MR experiences. As before, the \textit{id} parameter is automatically assigned to each HMD connecting to the architecture, thus ensuring that the system can scale up to accommodate multiple users, broadening the possible applications in multi-human-robot interaction scenarios.

\subsection{Kafka Component}

Apache Kafka is an open-source, high-performant, and distributed platform for data streaming based on the publish-subscribe paradigm. In the context of this work, it enables communication between the various components of our architecture while ensuring integration with other solutions, regardless of their usage of ROS.

The adoption of Kafka aims to overcome some inherent limitations of the publish-subscribe system for ROS and ROS2 architectures. On the one hand, ROS networking is based on the concept of a single master, with multiple nodes acting as producers and consumers. This configuration, however, has intrinsic scalability issues, as the master node can become the system's bottleneck in case of large throughput of data. In addition, such a scheme does not allow real-time constraints to be met since swarms of robots or highly complex robotic platforms may render the system unstable. On the other hand, ROS2 has been developed to solve the single master issue by adopting Data Distribution Service (DDS) middleware. Although this solution scales horizontally, providing a proper architecture for multiple robots, it still suffers from the limitation of a push-based architecture; a single server keeps track of all the subscribers and consumers and delivers the messages accordingly. For instance, in a robotics application where a robot is pushing information at a high frequency, this type of architecture is prone to bottlenecks because the server does not keep up with high messaging rates.

Conversely, by adopting Kafka the server is not responsible for sending messages to all consumers. Instead, it maintains an offset record, which consumers can request. Such a push/pull system provides horizontal scalability with a message replication feature, i.e., in the case of a faulty server, the messages will not be lost, and the performance will not be affected. We designed a \textit{ROS-Kafka interface} to extend the publish-subscribe capability of ROS with the distributed streaming service of Kafka. Each ROS node is treated as a master that can publish and subscribe to any topic, even those topics streamed by other master nodes, by the moment Kafka handles message sharing between the nodes. This approach not only provides a robust horizontally scalable solution, but also opens the possibility of integration with different robotics frameworks other than ROS, either in cooperation with ROS or as a full replacement.

\begin{figure*}[t!]
\centering
\begin{subfigure}{.46\textwidth}
  \centering
    \includegraphics[width=\linewidth]{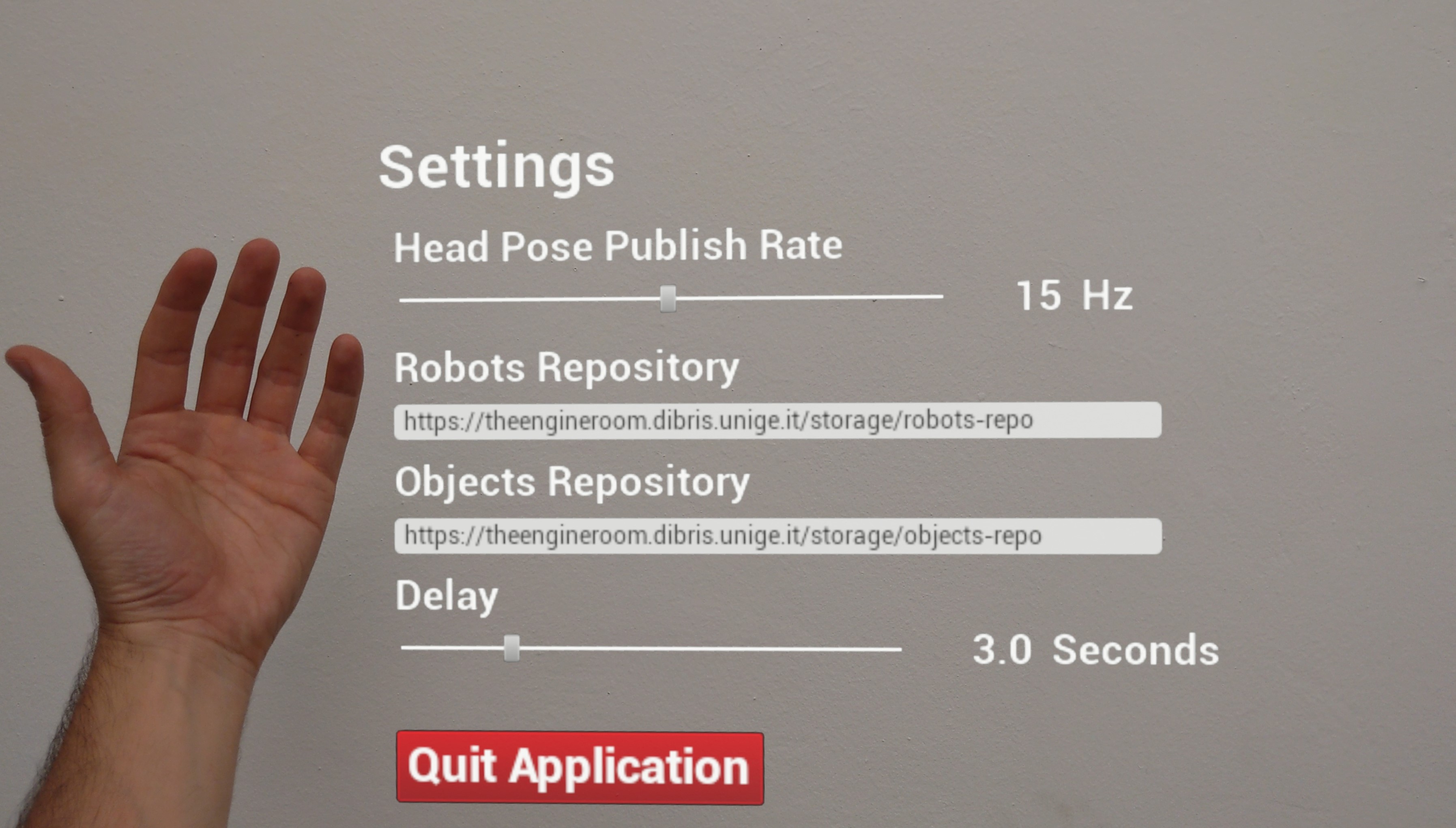}
    \caption{Settings menu.}
    \label{fig:settings-menu}
\end{subfigure}
\hfill
\begin{subfigure}{.46\textwidth}
  \centering
    \includegraphics[width=\linewidth]{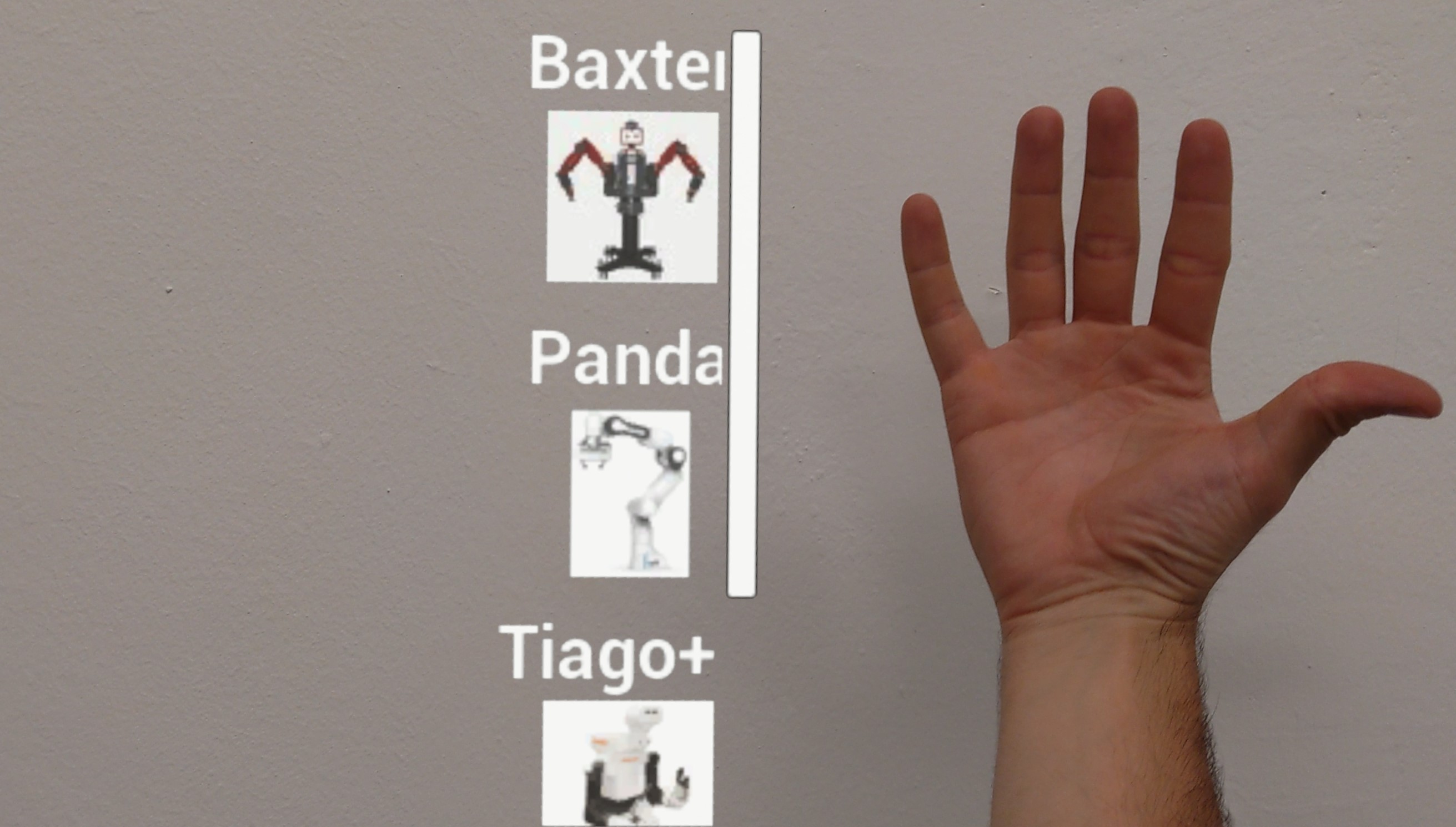}
    \caption{Model selection menu.}
    \label{fig:model-selection-menu}
\end{subfigure}
\caption{Overview of the holographic menus inside the MR layer, enabling users to customize the application's behavior.}
\label{fig:holo-menus}
\end{figure*}
 
Using a standard cloud-native protocol also facilitates the integration of the proposed architecture with Digital Twins (DTs), because it provides an abstracted communication level unrelated to the programming language or hardware used, thus rendering the system extendable. For instance, a DT running on a full-fledged PC or cloud server can perform heavy lifting and communicate visual feedback about the ongoing collaboration directly to our MR app running on the embedded HMD device.

\section{Example of Usage}
\label{sec:example-of-usage}

As already mentioned, the MR application can be employed off the shelf without further customization. 
It is sufficient to follow the steps in the GitHub repository to download its packaged version and install it on the HMD device, ideally a Microsoft HoloLens2, via a USB connection.

\subsection{Application Settings}

Upon launching the application, users can change settings by accessing the corresponding hand-attached menu, which is programmed to pop up whenever they look at their left hand. Fig. \ref{fig:settings-menu} depicts such settings menu as it appears to the user inside the holographic layer. As previously stated, the application ships with default links referencing public repositories made available with the application itself. Such repositories host, respectively, robots' and objects' resources. 
The former comprises assets (i.e., URDF and SDF files) associated with robot models commonly employed in research applications, including \textit{Baxter} from Rethink Robotics, \textit{TIAGo++} from Pal Robotics, \textit{Panda} from Franka Emika and \textit{UR5} from Universal Robots. On the other hand, the second repository stores models (i.e., FBX files) of simple and common items which can appear in HRC scenarios, including screwdrivers, hammers, and water bottles. Nevertheless, by simply modifying such links, users can point to their repositories, ensuring that other robot models and objects can be employed and loaded as holographic assets. It is important to note that the application can also deal with articulated objects, i.e., objects endowed with internal degrees of freedom. Such items, assuming they are properly described by a URDF file, can easily be spawned as holographic assets in the same way as robot models do.

Additionally, users can interact with the settings menu to customize the publication rate for the HMD's pose, which is triggered once mutual localization between the headset and QR code is established. Allowed rates range from a minimum frequency of $1$ \textit{Hz} up to a maximum of $30$ samples per second.

Finally, users can adjust the aforementioned delay controlling the elapsed time between holographic and subsequent robot actions. This feature makes it possible to experiment and find the optimal temporal distance between holographic intentions and actions, ensuring smooth interaction between agents and thus maximizing team efficiency.

\subsection{Spawning Robot Holograms}

To properly spawn robots inside the holographic scene, 
users should select the corresponding models via the menu attached to their right hand, which works in the same fashion as the settings. This menu, depicted in Fig. \ref{fig:model-selection-menu}, enables users to choose robots from a list of available resources. In particular, Fig. \ref{fig:model-selection-menu} shows the list with the four models described above, which ship with the default robots' remote repository. Once the custom repository is configured in the settings, on every startup the application will connect to it, checking if new robot models have been uploaded. In such cases, the corresponding resources are downloaded, thus making them available in the model selection menu for the user. Upon selecting the desired model, the user can close the menu, turn to the QR code and scan it with the HMD's camera, causing the robot hologram to spawn at the marker's estimated location. 
Subsequently, the user can manually adjust the holographic model's position by simply interacting with it through hand gestures if the robot's virtual replica and its physical counterpart are misaligned. From that moment on, the architecture exposes the topics associated with the newly spawned robot model, enabling the MR application to receive messages and display upcoming actions as holographic animations. 


\section{Conclusions} \label{sec:conclusions}

In this article, we introduced and detailed an open-source software architecture involving MR and HMD devices for conveying robots' intentions to human teammates in generalized contexts of HRC. The architecture has been designed with modularity and reusability principles in mind, providing off-the-shelf tools to other researchers working in the HRC field. As discussed throughout the work, the architecture can be used as-is in whatever scenario or integrated into more comprehensive solutions. 

Prior research studies have already demonstrated the effectiveness of the holographic communication modality implemented by the proposed architecture in collaborative assembly tasks involving a single human and their corresponding robotic teammate. Exploiting the intrinsic scalability features of the proposed work, future research studies could focus on developing shared and consistent MR experiences involving multiple robots and individuals in complex collaborative environments.

\bibliographystyle{IEEEtran}
\bibliography{bibliography}

\end{document}